\documentclass[mlmain]{jmlr}
\makeatletter
\def\@jmlrpages{}%
\makeatother

\usepackage{longtable}

\usepackage{booktabs}
\usepackage[load-configurations=version-1]{siunitx} 

\theorembodyfont{\upshape}
\theoremheaderfont{\scshape}
\theorempostheader{:}
\theoremsep{\newline}

\jmlrvolume{}
\firstpageno{1}

\title[Variational Manifold Embedding Framework]{A Variational Manifold Embedding Framework \\for Nonlinear Dimensionality Reduction}

\author{
\Name{John J. Vastola} \Email{john\_vastola@hms.harvard.edu}\\
\Name{Samuel J. Gershman} \Email{gershman@fas.harvard.edu}\\
\Name{Kanaka Rajan} \Email{kanaka\_rajan@hms.harvard.edu}\\
\addr Harvard University, Cambridge, MA, USA 
}

\begin{document}

\maketitle

\begin{abstract}
Dimensionality reduction algorithms like principal component analysis (PCA) are workhorses of machine learning and neuroscience, but each has well-known limitations. Variants of PCA are simple and interpretable, but not flexible enough to capture nonlinear data manifold structure. More flexible approaches have other problems: autoencoders are generally difficult to interpret, and graph-embedding-based methods can produce pathological distortions in manifold geometry. Motivated by these shortcomings, we propose a variational framework that casts dimensionality reduction algorithms as solutions to an optimal manifold embedding problem. By construction, this framework permits nonlinear embeddings, allowing its solutions to be more flexible than PCA. Moreover, the variational nature of the framework has useful consequences for interpretability: each solution satisfies a set of partial differential equations, and can be shown to reflect symmetries of the embedding objective. We discuss these features in detail and show that solutions can be analytically characterized in some cases. Interestingly, one special case exactly recovers PCA.

\end{abstract}

\section{Introduction}
\label{sec:intro}

Dimensionality reduction---the problem of identifying useful low-dimensional representations of high-dimensional data---is a central concern across machine learning, statistics, and neuroscience. But existing approaches have issues with accuracy, interpretability, or both. For example, principal component analysis (PCA) \citep{Pearson_PCA_1901,hotelling1933analysis,bishop_2006} is computationally cheap and can be interpreted in terms of a simple generative model \citep{tipping_probabilistic_1999}, but cannot capture nonlinear structure in data.

Two complementary perspectives have motivated lines of work that go beyond PCA. The \textit{geometric} perspective on dimensionality reduction casts the problem as identifying the low-dimensional structure (or \textit{data manifold}) along which data tend to lie. Methods like Locally Linear Embedding \citep{roweis_lle_2000}, Isomap \citep{tenenbaum_isomap_2000}, and diffusion maps \citep{coifman_diffusion_2006} exploit this perspective to learn a low-dimensional data manifold embedded in the high-dimensional ambient space. Work on so-called neural manifolds, which seeks low-dimensional representations of neural data, also tends to take this perspective \citep{cunningham_dimensionality_2014,chung_neural_2021,perich_neural_2025}. 

The \textit{probabilistic generative modeling} perspective on dimensionality reduction casts the problem as minimizing a divergence between the data distribution (or a suitably transformed version of it) and a distribution on a low-dimensional `latent' space. Neural-network-based autoencoders \citep{lecun1987phd,hinton_autoencoders_1993} and variational autoencoders \citep{kingma_VAE_2013,higgins2017betavae} exploit this perspective in order to capture arbitrarily complex (smooth) structure in data, but may yield results that are difficult to interpret. Methods which involve embedding samples into a graph, like Uniform Manifold Approximation and Projection (UMAP) \citep{umap2018} and t-distributed stochastic neighbor embedding (t-SNE) \citep{maaten_visualizing_2008}, are nonlinear and somewhat easier to interpret, but have been observed to produce pathologies in downstream statistical and clustering analyses \citep{wattenberg2016how,chari_specious_2023}. 

Motivated by these two perspectives, here we present a novel, nonlinear framework for dimensionality reduction that (i) involves minimizing a divergence between latent space distributions, like variational autoencoders; and (ii) constrains that optimization problem by making geometry-related \textit{nonparametric} assumptions about the mapping between latent and ambient space. We show that this framework has a variety of reasonable features, is amenable to mathematical analysis, and includes PCA as a special case. Critically, the variational structure of the problem permits us to use tools from physics to reason about properties of optimal embeddings, including that symmetries of the data distribution constrain them, and that they satisfy a certain set of partial differential equations (PDEs).

\section{A tractable variational framework for dimensionality reduction}
\label{sec:obj}

\paragraph{When are two data manifolds the same?} Let $\vec{x} \in \mathbb{R}^D$ denote an observation in a $D$-dimensional space, and $p_{\text{data}}(\vec{x})$ the associated data likelihood. When is $p_{\text{data}}$ more or less identical to some other likelihood $q$? Or, in more explicitly geometric language, when are the `data manifolds' associated with these likelihoods equivalent? One reasonable notion of equivalence comes from demanding that $q$ and $p_{\text{data}}$ can be obtained from one another via a smooth bijection (i.e., a \textit{diffeomorphism}) $\vec{g}: \mathbb{R}^D \to \mathbb{R}^D$. In particular, we ask that 
\begin{equation}
q(\vec{x}) \ d\vec{x} = p_{\text{data}}( \vec{g}(\vec{x}) ) \ |\det\vec{J}_{\vec{g}}(\vec{x})| \ d\vec{x} 
\end{equation}
almost everywhere, where $\vec{J}_{\vec{g}}(\vec{x})$ is the $D \times D$ Jacobian matrix whose entries are $J_{ij}(\vec{x}) := \partial g_i(\vec{x})/\partial x_j$. We will call the distribution $(\vec{g}^* p_{\text{data}})(\vec{x}) := p_{\text{data}}( \vec{g}(\vec{x}) ) \ |\det\vec{J}_{\vec{g}}(\vec{x})|$ the \textit{pullback} of $p_{\text{data}}$ by $\vec{g}$. Intuitively, the above equation says that one can obtain $p_{\text{data}}$ from $q$ by a smooth deformation of the state space $\mathbb{R}^D$. 

Dimensionality reduction involves attempting to capture \textit{most} of the structure of $p_{\text{data}}$ by relating it to a distribution defined on a generally lower-dimensional space $\mathbb{R}^d$ for some $d \leq D$. Let $\vec{z} \in \mathbb{R}^d$ denote an element of this `latent' space. Since it is usually impossible to capture \textit{all} high-dimensional structure of $p_{\text{data}}$ via a lower-dimensional approximation, we can only ask that $q(\vec{z}) d\vec{z}$ is \textit{approximately} the same as the pullback of $p_{\text{data}}$ through a smooth embedding map $\vec{\phi}: \mathbb{R}^d \to \mathbb{R}^D$ (which we denote $\vec{\phi}^* p_{\text{data}}$). More precisely, we want 
\begin{equation} \label{eq:goal_approx}
q(\vec{z}) \ d\vec{z} \approx p_{\text{data}}( \vec{\phi}(\vec{z}) ) \sqrt{\det( \vec{J}(\vec{z})^T \vec{J}(\vec{z}) )} \ d\vec{z} 
\end{equation}
almost everywhere, where $\vec{J}(\vec{z})$ is the $D \times d$ Jacobian with entries $J_{ij}(\vec{z}) := \partial \phi_i/\partial z_j$. Notice that, since we may have $d < D$, the change of variables correction involves the Riemannian volume form $\sqrt{\det(\vec{J}(\vec{z})^T \vec{J}(\vec{z}) )}$ (we cannot use $|\det\vec{J}(\vec{z})|$, since it may equal zero).

\begin{figure}[t]
\floatconts
  {fig:first}
  {\caption{\textbf{a.} The function $\vec{\phi}$ maps the latent space to the (higher-dimensional) ambient space. \textbf{b.} Optimal one-dimensional embeddings when $p_{\text{data}}$ is a mixture of Gaussians centered at points along a line (left), circle (middle), or sinusoid (right).}}
  {\includegraphics[width=\linewidth]{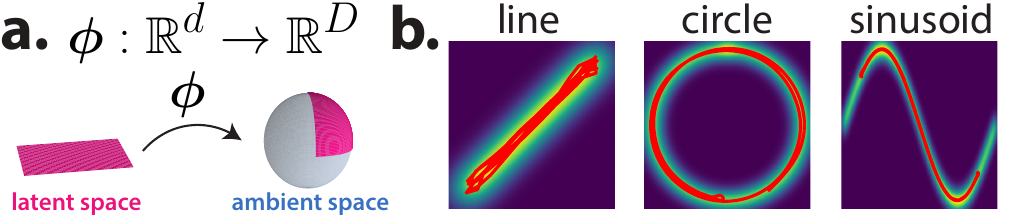}}
\end{figure}

\paragraph{A variational objective for dimensionality reduction.} Let $\vec{\phi}: \mathbb{R}^d \to \mathbb{R}^D$ be a smooth injection from a `latent' space to an `ambient' space (with $d \leq D$), and suppose that $\vec{z} \in \mathbb{R}^d$ is distributed according to a \textit{prior} $q(\vec{z})$, and $\vec{x} \in \mathbb{R}^D$ is distributed according to a likelihood $p_{\text{data}}$. Motivated by Eq. \ref{eq:goal_approx}, we define an \textit{optimal dimensionality reduction algorithm} (or \textit{optimal embedding}) as a map $\vec{\phi}$ which \textit{maximizes}
\begin{equation} \label{eq:J_basic}
\boxed{ \ J[\vec{\phi}] = \int_{\mathbb{R}^d} d\vec{z} \ q(\vec{z}) \left[ \frac{1}{2} \log\det(\vec{J}(\vec{z})^T \vec{J}(\vec{z}) ) + \log p_{\text{data}}(\vec{\phi}(\vec{z})) - \log q(\vec{z})   \right]  \ } \ ,
\end{equation}
i.e., the negative of the Kullback-Leibler (KL) divergence between $q$ and $\vec{\phi}^* p_{\text{data}}$, while satisfying the finiteness conditions (see Appendix \ref{apd:obj_basics})
\begin{equation} \label{eq:J_bc}
 \mathbb{E}_{\vec{z} \sim q}\left\{ \sqrt{\det(\vec{J}(\vec{z})^T \vec{J}(\vec{z}) )} \right\} < \infty  \hspace{0.2in} \text{ and } \hspace{0.2in}  \mathbb{E}_{\vec{z} \sim q}\left\{ \Vert \vec{\phi}(\vec{z}) \Vert_2^2 \right\} < \infty  \ .
\end{equation}
Because $J = - D_{\text{KL}}( q \Vert \vec{\phi}^* p_{\text{data}})$, we immediately know that $J$ is upper bounded by zero,\footnote{Note that this is only true if $\vec{\phi}$ is an injection. See Appendix \ref{apd:obj_basics}.} which corresponds to perfect performance. See \figureref{fig:first}a for a schematic.

This objective has two terms, each of which perform slightly different roles. The log-likelihood term encourages $\vec{\phi}$ to map into regions of high likelihood, while the log-determinant term encourages embeddings to have a nontrivial volume. (If only the log-likelihood term were present, an optimal embedding may map all $\vec{z}$ near a global maximum of $p_{\text{data}}$.) Interestingly, these terms also have clear physical analogues: the log-likelihood term functions like a \textit{potential energy}, and the log-determinant term, which depends on derivatives of the embedding, functions like a \textit{kinetic energy}.

As in physics, the fact that the objective depends not just on the embedding coordinates $\phi_i(\vec{z})$, but also on the derivatives $J_{ij} := \partial \phi_i/\partial z_j$, allows us to directly apply results from the calculus of variations \citep{goldstein1980mechanics,elsgolc2007calculus} that cannot be usefully leveraged in more generic (e.g., autoencoder) settings. For more discussion of this objective, including a Bayesian interpretation of it that casts optimizing $J$ as identifying the maximum a posteriori estimate of $\vec{\phi}$, see Appendix \ref{apd:obj_basics}.

\paragraph{Examples of optimal one-dimensional embeddings.} What do optimal embeddings look like? The naive way to estimate them is to parameterize $\vec{\phi}$ and numerically optimize $J$ via a gradient-descent-like algorithm. If we parameterize $\vec{\phi}$ as a multilayer perceptron and choose $p_{\text{data}}$ to be different Gaussian mixtures, optimizing $J$ in the standard way tells us that optimal \textit{one-dimensional} embeddings are generally nonlinear, and correspond to what one might expect: if there is an obvious path through the points, the optimal embedding follows it (\figureref{fig:first}b; code: \url{https://github.com/john-vastola/varembed-neurreps25}).

\section{Interpreting optimal one-dimensional embeddings}
\label{sec:1D}

We can say more about optimal one-dimensional embeddings, and optimal embeddings more generally, by exploiting results from the calculus of variations. The most basic such result is that optimal embeddings satisfy the \textit{Euler-Lagrange} (EL) \textit{equations}, which we define next.

\subsection{Optimal one-dimensional embeddings follow score vectors}

If we approximate $p_{\text{data}}$ with a one-dimensional manifold ($d = 1$), the objective becomes
\begin{equation} \label{eq:J_basic_1D}
 J[\vec{\phi}] = \int_{-\infty}^{\infty} dz \ q(z) \left[ \frac{1}{2} \log \Vert \vec{\phi}'(z) \Vert_2^2 + \log p_{\text{data}}(\vec{\phi}(z)) - \log q(z)   \right] 
\end{equation}
where $\vec{\phi}'(z) := ( \phi'_1(z), ..., \phi'_D(z) )^T$ is the derivative of the embedding. This version of the objective looks extremely similar to a classical mechanics problem, with the main qualitative difference being that the kinetic term is $\frac{1}{2} \log \Vert \vec{\phi}'(z) \Vert_2^2$ instead of $\frac{1}{2} \Vert \vec{\phi}'(z) \Vert_2^2$.

We can view $\vec{\phi}$ as tracing out a one-dimensional trajectory through $\mathbb{R}^D$. According to a standard result in the calculus of variations, these trajectories satisfy the EL equations
\begin{equation}
\frac{d}{dz} \left( \frac{\partial L}{\partial \phi'_i} \right) = \frac{\partial L}{\partial \phi_i} \hspace{0.2in} \text{, i.e.,} \hspace{0.2in} \frac{d}{dz} \left[  q(z) \frac{\phi_i'(z)}{\Vert \vec{\phi}'(z) \Vert_2^2} \right] = q(z) \frac{\partial}{\partial \phi_i} \log p_{\text{data}}(\vec{\phi}(z)) 
\end{equation}
for each $i = 1, ..., D$ \citep{goldstein1980mechanics,elsgolc2007calculus}, where the \textit{Lagrangian} $L$ is defined as
\begin{equation}
L( \vec{\phi}, \vec{\phi}', z) := q(z) \left[ \frac{1}{2} \log \Vert \vec{\phi}'(z) \Vert_2^2 + \log p_{\text{data}}(\vec{\phi}(z)) - \log q(z)   \right] \ .
\end{equation}
These equations place nontrivial constraints on the form of optimal embeddings; satisfying them is necessary, but not sufficient, for $\vec{\phi}$ to be a global maximizer of $J$. If we define momentum-like variables $p_i(z) := \phi_i'(z)/\Vert \vec{\phi}'(z) \Vert_2^2$, we find that the `dynamics' defined by the EL equations looks like that of a massive particle moving through the landscape determined by the potential $- \log p_{\text{data}}$:
\begin{equation} \label{eq:mom_dyn}
\begin{split}
\phi_i'(z) = \frac{p_i(z)}{\Vert \vec{p}(z) \Vert_2^2} \hspace{0.4in} p_i'(z) = s_i(\vec{\phi}(z)) - \left[ \frac{d}{dz} \log q(z) \right] p_i(z) 
\end{split}
\end{equation}
where $s_i(\vec{\phi}(z)) := \frac{\partial}{\partial \phi_i} \log p_{\text{data}}(\vec{\phi}(z))$ is the \textit{score vector} associated with $p_{\text{data}}$ at the point $\vec{\phi}(z) \in \mathbb{R}^D$. These vectors, which play a foundational role in how diffusion models sample \citep{song2021scorebased} and generalize \citep{wang2024the,vastola2025generalization}, point in the direction in which the data likelihood increases most. Eq. \ref{eq:mom_dyn} says that the trajectory $\vec{\phi}(z)$ moves in the direction of the score $\vec{s}(\vec{\phi}(\vec{z}))$, and that the influence of the score vector is controlled by a $q$-dependent `friction' factor and the norm of $\vec{p}(z)$. The norm $\Vert \vec{p}(z) \Vert_2^2$ plays the role of a `mass'. When $\vec{p}(z)$ is large (which can happen when the score vectors become large), the trajectory changes very slowly; when it is small, the trajectory changes quickly.

\subsection{Connection to diffusion models}

We can make the connection between optimal one-dimensional embeddings and diffusion models more concrete by observing that one significant difference between Eq. \ref{eq:mom_dyn} and a probability flow ODE \citep{song2021scorebased} is that the former involves \textit{inertia}: $\vec{\phi}(z)$ does not \textit{only} move in the direction of the score vector. Since the amount of inertia that appears in Eq. \ref{eq:mom_dyn} depends on the prior $q$, we can modulate it by choosing a special prior. If we pick $q(z) = \gamma e^{\gamma z}$ for $z \in (-\infty, 0]$, in the large $\gamma$ limit we obtain dynamics (see Appendix \ref{apd:emb_1d})
\begin{equation} 
\phi_i'(z) = \gamma \frac{s_i(\vec{\phi}(z))}{\Vert \vec{s}(\vec{\phi}(z)) \Vert_2^2} 
\end{equation}
which follows score vectors, but moves faster or slower depending on their magnitude. If we reparameterize `time', we recover an ODE that follows score vectors without the norm-dependent distortion, and hence looks like a probability flow ODE at a fixed noise scale.

\section{Useful properties of the variational objective}
\label{sec:useful_props}

What other properties of optimal embeddings can we reason about without actually optimizing the objective? In this section, we discuss two kinds: properties related to \textit{flexibility}, and properties related to \textit{interpretability}. Unlike in the previous section, here our comments apply for any choice of latent dimensionality (i.e., any $d \leq D$).

\subsection{Flexibility-related properties}

\paragraph{Objective is geometry-sensitive but not coordinate-sensitive.} Our framework ought to reflect the geometry of the data manifold and prior, but not our incidental choice of coordinates;\footnote{In this respect, embedding is similar to other geometry-sensitive problems, like close-packing \citep{Viazovska2017,vastola_packing_2024}.} in particular, if we relabeled each $\vec{z} \in \mathbb{R}^d$ according to a diffeomorphism $g: \mathbb{R}^d \to \mathbb{R}^d$, up to this relabeling, we would like $J$ to have the same value and solutions. This reparameterization-invariance property trivially holds, since $J$ equals a (negative) KL divergence, and the property holds for KL divergences. It is also straightforward to show that this property holds through direct computation (see Appendix \ref{apd:basic_props}).

\paragraph{Objective has expected trivial solutions.} Because $J$ is bounded from above by zero, we know that any choice of $\vec{\phi}$ that makes it achieve this bound is a global optimum. This property makes it clear that $J$ has two types of reasonable trivial solutions. First, when $d = D$ and the prior $q$ equals the data likelihood $p_{\text{data}}$---i.e., when there is no dimensionality reduction---the identity map $\vec{\phi}(\vec{z}) = \vec{z}$ is optimal since it makes $J$ equal zero. Second, if $p_{\text{data}}$ is genuinely low-dimensional (say, $d$-dimensional), and there exists a smooth injection $\vec{\psi}$ for which the pullback $\vec{\psi}^* p_{\text{data}}$ is normalized on $\mathbb{R}^d$, then $\vec{\phi} = \vec{\psi}$ is a global optimum. Importantly, this is true even if $\vec{\psi}$ is highly nonlinear.

\subsection{Interpretability-related properties}

\paragraph{Optimal embeddings satisfy EL equations.}

The Lagrangian density corresponding to our objective (i.e., its integrand) is
\begin{equation}
\mathcal{L}( \vec{\phi}, \vec{J}, \vec{z}) := q(\vec{z}) \left[ \frac{1}{2} \log\det(\vec{J}^T \vec{J}) + \log p_{\text{data}}(\vec{\phi}(\vec{z})) - \log q(\vec{z})  \right] \ .
\end{equation}
A useful consequence of our variational formulation is that it allows us to exploit mathematical results from the calculus of variations. In particular, optimal mappings $\vec{\phi}$ satisfy the Euler-Lagrange (EL) equations (see Appendix \ref{apd:EL})
\begin{equation}
\sum_{j = 1}^d \partial_j \left( \frac{\partial \mathcal{L}}{\partial (\partial_j \phi_i)} \right) = \frac{\partial \mathcal{L}}{\partial \phi_i}  \ \text{, i.e., } \ \sum_j \left[ \partial_j ( J_{ji}^+ ) + J_{ji}^+ \partial_j [ \log q(\vec{z}) ] \right] = \frac{\partial \log p_{\text{data}}(\vec{\phi}(\vec{z}))}{\partial \phi_i} \ ,
\end{equation}
where $\vec{J}^+ := (\vec{J}^T \vec{J})^{-1} \vec{J}^T \in \mathbb{R}^{d \times D}$ is the Moore-Penrose pseudoinverse of $\vec{J}$ (which satisfies $\vec{J}^+ \vec{J} = \vec{I}_d$), and where we have relied on a well-known result on the derivative of a log-determinant \citep{cookbook}. 

The EL equations define a system of $D$ coupled nonlinear PDEs: there is one equation for each $\phi_i(\vec{z})$, and only up to second derivatives of the $\phi_k$ appear. (Unfortunately, due to the presence of $\vec{J}^+$, one may encounter derivatives of the $\phi_k$ in denominators.) Satisfying the EL equations is necessary but not sufficient for $\vec{\phi}$ to maximize $J$, which allows one to use these PDEs to strongly constrain the solution space.

\paragraph{Continuous symmetries of the objective constrain solutions.} We expect that symmetries of the data likelihood constrain how one ought to perform dimensionality reduction. For example, the dimensionality-reduced version of a radially symmetric likelihood is also probably radially symmetric. \textit{Noether's theorem} \citep{Noether1918,Neuenschwander2017-wonderful,V2025noether}, a powerful result frequently used in field theory settings, makes this intuition precise. It states that if a small perturbation $\vec{\phi}(\vec{z}) \to \vec{\phi}(\vec{z}) + \delta \vec{\phi}(\vec{z})$ of the embedding causes the Lagrangian density to change by a total derivative, i.e., $\mathcal{L} \to \mathcal{L} + \nabla \cdot \vec{K}$ for some vector $\vec{K}$, then the quantity (the \textit{Noether current})
\begin{equation}
j_a := \sum_{b = 1}^D \frac{\partial \mathcal{L}}{\partial (\partial_{a} \phi_b)} (\delta \phi_{b}) - K_a
\end{equation}
is conserved (i.e., $\nabla \cdot \vec{j} = 0$) when the EL equations are satisfied (see Appendix \ref{apd:noether}). Noether's theorem is useful because it allows us to make concrete statements about how continuous symmetries constrain optimal embeddings, even if and especially when the EL equations are too difficult to solve analytically.

\section{Conserved quantities and consequences of continuous symmetries}
\label{sec:noether}

In this section, we study explicit examples of how continuous symmetries constrain optimal embeddings. We consider three kinds of continuous symmetry: \textit{reparameterization invariance}, which yields an analogue of energy conservation; \textit{embedding translation invariance}, which yields an analogue of momentum conservation; and \textit{embedding rotation invariance}, which yields an analogue of angular momentum conservation.

\subsection{Embedding energy is conserved for optimal embeddings}


If $q$ is uniform, the Lagrangian density is locally reparameterization invariant, since an infinitesimal change of coordinates changes $\mathcal{L}$ by a total derivative. By Noether's theorem, this implies that the \textit{stress-energy tensor}
\begin{equation}
T_{i j} = \sum_k \frac{\partial \mathcal{L}}{\partial (\partial_i \phi_k)} (\partial_j \phi_k(\vec{z})) - \delta_{i j} \mathcal{L}(\vec{\phi}, \vec{J}, \vec{z}) = \sum_k J^+_{i k} J_{k j} q - \delta_{ij} \mathcal{L}(\vec{\phi}, \vec{J}, \vec{z}) = \delta_{i j} ( q - \mathcal{L} )
\end{equation}
is conserved ($\sum_i \partial_i T_{ij} = 0$). But since $T_{ij}$ is only nonzero when $i = j$, conservation just means that $\mathcal{L} = \text{const.}$ Hence, $\mathcal{L}$ being conserved is analogous to the conservation of energy.

Usefully, this result can be combined with the reparameterization invariance of the objective to establish a more general result that holds even for non-uniform $q$ (one can also derive this result through a direct computation; see Appendix \ref{apd:noether}): the \textbf{embedding energy}
\begin{equation}
\boxed{ \ \mathcal{E} := -\frac{1}{2} \log\det(\vec{J}^T \vec{J}) - \log p_{\text{data}}(\vec{\phi}(\vec{z})) + \log q(\vec{z}) \ } 
\end{equation}
is constant (i.e., does not depend on $\vec{z}$) for all solutions to the EL equations (and in particular, for the global maximizers of $J$). Interestingly, since
\begin{equation}
0 \geq J[\vec{\phi}] = - \int_{\mathbb{R}^d} d\vec{z} \ q(\vec{z}) \ \mathcal{E} = - \mathcal{E} \ ,
\end{equation}
embedding energy is always nonnegative, and global maximizers of $J$ \textit{minimize} it.

\subsection{Translation invariance yields momentum conservation analogue}

If $p_{\text{data}}$ is uniform along some ambient space direction, so that it does not depend on $\phi_i$, Noether's theorem implies that the \textit{canonical momentum} along that direction is conserved:
\begin{equation}
\sum_j \partial_j \left( \frac{\partial \mathcal{L}}{\partial (\partial_j \phi_i)} q(\vec{z}) \right) = \sum_j \partial_j\left[ J_{j i}^+ q(\vec{z}) \right]  = 0 \ .
\end{equation}
This conservation law almost fully characterizes the optimal embedding when $p_{\text{data}}$ is uniform (on a ball of radius one-half centered at $\vec{x} = (1/2, ..., 1/2)^T$, say) and $d = 1$. Together with embedding energy conservation, it implies $\phi'_i(z) = \text{const.}/q(\vec{z})$, so
\begin{equation}
\phi_i(z) =  \int_{-\infty}^z q(y) \ dy 
\end{equation}
for all $i = 1, ..., D$. Interestingly, this mapping exactly matches the one used by efficient codes to convert a generally non-uniform stimulus distribution to a uniform distribution.

\subsection{Rotation invariance yields angular momentum conservation analogue}

If $p_{\text{data}}$ is invariant to rotations in the $\phi_i$-$\phi_j$ plane, Noether's theorem implies that 
\begin{equation}
\sum_a \partial_a L^a_{ij} = 0 \hspace{0.2in} \text{ where } \hspace{0.2in} L^a_{i j} := q(\vec{z})  \left[  J^+_{a i} \phi_j - J^+_{a j} \phi_i   \right]  \ .
\end{equation}
When $d = 1$, $\vec{J} \to \vec{\phi}'$ and $\vec{J}^+ \to \vec{\phi}'/\Vert \vec{\phi}'  \Vert_2^2$, so angular momentum conservation becomes
\begin{equation}
q(z) \frac{[ \phi_1'(z) \phi_2(z) - \phi_1(z) \phi_2'(z) ]}{\Vert \vec{\phi}'(z) \Vert_2^2} = \text{const.}
\end{equation}
When combined with embedding energy conservation, this conservation law can be used to show that optimal embeddings lie along circles when $D = 2$. Hence, the optimal embedding for a ring attractor structure, as expected, follows the ring.

\section{A solvable case: PCA is optimal for Gaussian prior and likelihood}
\label{sec:main_pca}

Optimal embeddings are difficult to determine analytically, in part because the EL equations are generally a system of highly nonlinear coupled PDEs. Do there exist any interesting solvable cases? Perhaps surprisingly, the answer is yes: the optimal embedding exactly corresponds to PCA when (i) the prior is an isotropic normal distribution $q(\vec{z}) = \mathcal{N}(\vec{z}; \vec{0}, \sigma_0^2 \vec{I})$, with $\sigma_0 > 0$; and (ii) the likelihood is a multivariate Gaussian $p_{\text{data}}(\vec{x}) = \mathcal{N}(\vec{x}; \vec{0}, \vec{\Sigma})$, with $\vec{\Sigma}$ a positive definite $D \times D$ covariance matrix. 

The PCA solution corresponds to the linear mapping $\vec{\phi}(\vec{z}) = \sum_{k = 1}^d \frac{\sqrt{\lambda_k}}{\sigma_0} (\vec{q}_k^T \vec{z}) \vec{v}_k $, where the $\vec{v}_k$ are the top $d$ (normalized) eigenvectors of $\vec{\Sigma}$, and $\{ \vec{q}_k \}_{k = 1, ..., d}$ is any set of orthonormal vectors (i.e., $\vec{q}_k^T \vec{q}_{\ell} = \delta_{k \ell}$). Although it is not mathematically obvious that the EL equations admit such a simple solution, the intuition is fairly clear: one can obtain an anisotropic Gaussian from an isotropic one by uniformly stretching it in different directions.

\paragraph{The one-dimensional case.} Let us begin with the relatively simple $d = 1$ case, with
\begin{equation} \label{eq:J_pca1_main}
J[\vec{\phi}] := \int_{-\infty}^{\infty} dz \ \frac{e^{- \frac{1}{2 \sigma_0^2} z^2}}{\sqrt{2 \pi \sigma_0^2}} \ \left\{  \frac{1}{2} \log \Vert \vec{\phi}'(z) \Vert_2^2 - \frac{1}{2} \vec{\phi}(z)^T \vec{\Sigma}^{-1} \vec{\phi}(z)    \right\} + \text{const.}
\end{equation}
If the optimal $\vec{\phi}$ \textit{does} correspond to PCA, it should pick out the first principal component, i.e., an eigenvector of $\vec{\Sigma}$ whose eigenvalue is largest.\footnote{If this is not unique, we expect multiple solutions.} The EL equations read
\begin{equation}
\frac{d}{dz} \left[  q(z) \frac{\vec{\phi}'(z)}{\Vert \vec{\phi}'(z) \Vert_2^2} \right] = - q(z) \vec{\Sigma}^{-1} \vec{\phi}(z) \ .
\end{equation}
Consider a linear ansatz $\vec{\phi}(z) = \vec{v} z$ for some vector $\vec{v} \in \mathbb{R}^D$. Since $\vec{\phi}'(z) = \vec{v}$, we require
 \begin{equation}
\frac{d}{dz} \left[  q(z) \frac{\vec{v}}{\Vert \vec{v} \Vert_2^2} \right] = - q(z) \vec{\Sigma}^{-1} \vec{v} z \hspace{0.2in} \implies \hspace{0.2in} \vec{\Sigma} \vec{v} = (\sigma_0^2 \Vert \vec{v} \Vert_2^2) \ \vec{v} 
\end{equation}
i.e., that $\vec{v}$ must be an eigenvector of $\vec{\Sigma}$ with eigenvalue $\lambda = (\sigma_0^2 \Vert \vec{v} \Vert_2^2)$. Hence, the EL equations force $\vec{v}$ to be \textit{one} of the eigenvectors of $\vec{\Sigma}$, but not any specific eigenvector. The symmetry is broken by asking that $\vec{\phi}$ maximize $J$ (Eq. \ref{eq:J_pca1_main}). Its $\vec{\phi}$-dependent terms equal
\begin{equation}
\mathbb{E}_{z \sim q}\left\{ - \frac{z^2}{2}  \vec{v}^T \vec{\Sigma}^{-1} \vec{v} + \frac{1}{2} \log(\vec{v}^T \vec{v})   \right\} =  - \frac{\sigma_0^2}{2}  \vec{v}^T \vec{\Sigma}^{-1} \vec{v} + \frac{1}{2} \log(\vec{v}^T \vec{v}) = - \frac{1}{2} + \frac{1}{2} \log\left( \frac{\lambda}{\sigma_0^2} \right) 
\end{equation}
given our ansatz, which is maximized when $\lambda$ is as large as possible. This happens precisely when $\vec{v}$ is the top eigenvector of $\vec{\Sigma}$, which (almost) finishes our argument. Do the EL equations admit nonlinear solutions? We can show the answer is no by exploiting embedding energy conservation (see Appendix \ref{apd:pca_nonlinear}), which means $\vec{\phi}(z) = z\vec{v}$ is indeed optimal.

\paragraph{The general case.} For arbitrary $d \leq D$, the objective is
\begin{equation}
J[\vec{\phi}] := \int_{\mathbb{R}^d} d\vec{z} \ \frac{e^{- \frac{1}{2 \sigma_0^2} \Vert \vec{z} \Vert_2^2}}{(2 \pi \sigma_0^2)^{d/2}} \ \left\{ \frac{1}{2} \log\det[ \vec{J}(\vec{z})^T \vec{J}(\vec{z})] - \frac{1}{2} \vec{\phi}(\vec{z})^T \vec{\Sigma}^{-1} \vec{\phi}(\vec{z})  \right\} + \text{const.}
\end{equation}
Although the EL equations are more complicated than in the one-dimensional case, if we assume a linear solution $\vec{\phi}(\vec{z}) = \vec{A} \vec{z}$ for some $D \times d$ matrix $\vec{A}$, we recover an analogous constraint:
\begin{equation}
\vec{\Sigma} \vec{A} = \sigma_0^2 \vec{A} \vec{A}^T \vec{A} \ .
\end{equation}
If we exploit a singular value decomposition $\vec{A} = \sum_{k = 1}^d s_k \vec{u}_k \vec{q}_k^T$ of $\vec{A}$, where $\vec{u}_k^T \vec{u}_{\ell} = \vec{q}_k^T \vec{q}_{\ell} = \delta_{k \ell}$ and the $s_k$ are the singular values of $\vec{A}$, then this condition becomes
\begin{equation}
\vec{\Sigma} \vec{u}_k = (\sigma_0^2 s_k^2) \vec{u}_k \ .
\end{equation}
for all $k = 1, ..., d$. As before, this implies each $\vec{u}_k$ is an eigenvector of $\vec{\Sigma}$---although they must be \textit{different} eigenvectors, since they are pairwise orthogonal. Also as before, the symmetry is broken by asking that $\vec{\phi}$ maximize $J$. The $\vec{\phi}$-dependent part of $J$ equals
\begin{equation}
\mathbb{E}_{\vec{z} \sim q}\left\{ - \frac{1}{2}  \vec{z}^T \vec{A}^T \vec{\Sigma}^{-1} \vec{A} \vec{z} + \frac{1}{2} \log \det (\vec{A}^T \vec{A})   \right\} =  - \frac{d}{2} + \frac{1}{2} \sum_{k = 1}^d \log\left( \frac{\lambda_k}{\sigma_0^2} \right) 
\end{equation}
given our ansatz, which is maximized when each $\lambda_k$ is chosen to be as large as possible. This happens precisely when the $\{ \vec{u}_k \}$ are chosen to be the top $d$ eigenvectors of $\vec{\Sigma}$. Finally, we can argue that the solutions of the EL equations \textit{must} be linear (see Appendix \ref{apd:pca_nonlinear}).

\section{Discussion}

We introduced a variational framework for dimensionality reduction that is more flexible than PCA---and which includes it as a special case---and which has various features that aid interpretability, including that solutions satisfy PDEs, and that Noether's theorem links data manifold symmetries to conservation laws. Since each dimensionality reduction algorithm associated with the framework optimizes the objective for specific choices of $d$, $q$ and $p_{\text{data}}$, we can use these quantities to reason about the inductive biases of solutions. Our work is in the spirit of recent studies that leverage insights from physics to clarify the structure of machine learning algorithms \citep{Tanaka2021-noether-learning,vastola2025gradient}.

\paragraph{Limitations.} At least in the framework's current form, it assumes that both $q$ and $p_{\text{data}}$ are distributions on continuous rather than discrete spaces, making it impossible to directly apply it to (for example) Poisson-like spiking data. On the other hand, since distributions of spikes are plausibly related by Bayes' rule to the distributions of continuous latent variables (e.g., an animal's allocentric position in space), an intuition made precise in encoding models like generalized linear models \citep{weber_capturing_2017} and probabilistic population codes \citep{ma_bayesian_2006,vastola_ppc2023}, there is still a sense in which the framework can be applied indirectly. Also, while the framework could in principle be used to approximate data manifolds of any dimensionality, it is unclear how scalable it is, and how useful certain analytic insights (e.g., related to the EL equations) are for high-dimensional data sets of interest. Symmetries, for example, strongly constrain low-dimensional embeddings, but not necessarily high-dimensional ones.

\begin{acks}
SJG was funded by the Kempner Institute for the Study of Natural and Artificial Intelligence, and a Polymath Award from Schmidt Sciences. KR was funded by the NIH (RF1DA056403, U01NS136507), James S. McDonnell Foundation (220020466), Simons Foundation (Pilot Extension-00003332-02), McKnight Endowment Fund, CIFAR Azrieli Global Scholar Program, NSF (2046583), a Harvard Medical School Neurobiology Lefler Small Grant Award, and a Harvard Medical School Dean’s Innovation Award. This work has been made possible in part by a gift from the Chan Zuckerberg Initiative Foundation to establish the Kempner Institute for the Study of Natural and Artificial Intelligence at Harvard University.
\end{acks}

\clearpage

\bibliography{pca-bib}

\clearpage

\appendix

\section{Motivating the objective function}
\label{apd:obj_basics}

In this appendix, we present an alternative Bayesian argument to motivate our objective function
\begin{equation}  \label{eq:J_app}
 J[\vec{\phi}] = \int_{\mathbb{R}^d} d\vec{z} \ q(\vec{z}) \left[ \frac{1}{2} \log\det(\vec{J}(\vec{z})^T \vec{J}(\vec{z}) ) + \log p_{\text{data}}(\vec{\phi}(\vec{z})) - \log q(\vec{z})   \right]    \ ,
\end{equation}
and also briefly comment on technical subtleties related to this objective. As described in Sec. \ref{sec:obj}, $D$ is the ambient space dimensionality, $d \leq D$ is the latent space dimensionality, $p_{\text{data}}$ is a likelihood on $\mathbb{R}^D$, $q$ is a distribution (the `prior') on $\mathbb{R}^d$, and $\vec{\phi}: \mathbb{R}^d \to \mathbb{R}^D$ is a smooth injection chosen to maximize $J$, subject to certain finiteness conditions.

\subsection{A Bayesian argument for the objective function}

\paragraph{Assumed ground truth distribution.} Let $\vec{x} \in \mathbb{R}^D$ denote an observation. Assume that observations genuinely have low-dimensional structure, in the sense that they can be viewed as generally lying on or near some $d$-dimensional manifold embedded in $\mathbb{R}^D$ (with $d \leq D$). 

We can make the idea of a low-dimensional manifold more concrete by imagining that there exists some smooth injection $\vec{\phi}: \mathbb{R}^d \to \mathbb{R}^D$ which embeds any particular location $\vec{z} \in \mathbb{R}^d$ on this manifold (given in coordinates that may differ from those of the ambient space) into $\mathbb{R}^D$. We can make our demand that observations generally lie `near' the manifold more concrete by assuming a noise model $p_{\text{noise}}(\vec{x} | \vec{z}, \vec{\phi}) := \mathcal{N}(\vec{x}; \vec{\phi}(\vec{z}), \epsilon \vec{I}_D)$, where $\epsilon > 0$ is small. Note that this noise model allows observations to be off-manifold, but that this only occurs with a low probability.

If we assume that latent states are sampled according to a distribution $q(\vec{z})$, then the joint distribution of observations and latent states is
\begin{equation}
p(\vec{x}, \vec{z}) := p_{\text{noise}}(\vec{x} | \vec{z}, \vec{\phi}) q(\vec{z}) = \mathcal{N}(\vec{x}; \vec{\phi}(\vec{z}), \epsilon \vec{I}_D) q(\vec{z}) \ .
\end{equation}
Assume that $q$ and $\epsilon > 0$ are known, but the mapping $\vec{\phi}$ is not. How can we infer it?

\paragraph{Assumed model.} One sensible way to proceed is to exploit Bayes' rule. If we model observations as distributed according to $p_{\text{data}}(\vec{x})$, our knowledge of the noise model $p_{\text{noise}}(\vec{x} | \vec{z}, \vec{\phi})$ allows us to infer latent states from observations via
\begin{equation}
p(\vec{z} | \vec{\phi}) := \int p(\vec{z} | \vec{x}, \vec{\phi}) p_{\text{data}}(\vec{x}) \ d\vec{x} = \int \frac{p_{\text{noise}}(\vec{x} | \vec{z}, \vec{\phi}) q(\vec{z})}{\int p_{\text{noise}}(\vec{x} | \vec{z}', \vec{\phi}) q(\vec{z}') d\vec{z}'} p_{\text{data}}(\vec{x}) \ d\vec{x} \ .
\end{equation}
Let $p(\vec{\phi})$ denote a prior over mappings that encodes our preferences about it (e.g., that it is continuous). By Bayes' rule, the posterior distribution of $\vec{\phi}$ given a latent state $\vec{z}$ is
\begin{equation}
p(\vec{\phi} | \vec{z}) = \frac{p(\vec{z} | \vec{\phi}) p(\vec{\phi})}{\int p(\vec{z} | \vec{\phi}') p(\vec{\phi}') \ d\vec{\phi}'} \ ,
\end{equation}
where we note that, since the distributions over $\vec{\phi}$ are defined over \textit{function spaces}, there are technical subtleties related to normalization. Fortunately, these difficulties are irrelevant from the point of view of identifying the most likely mapping $\vec{\phi}$. 

\paragraph{Using Bayes' rule to estimate the embedding map.}  Let $\{ \vec{z}_i \}_{i = 1}^N$ be $N \geq 1$ independent and identically distributed samples from $q$, the ground truth distribution of latent states. We can estimate $\vec{\phi}$ by maximizing the scaled log-posterior probability
\begin{equation}
\begin{split}
 \frac{1}{N} \log p(\vec{\phi} | \{ \vec{z}_i \}) &= \frac{1}{N} \log p(\{ \vec{z}_i \} | \vec{\phi}) + \frac{1}{N} \log p(\vec{\phi}) + \text{const.} \\
&= \frac{1}{N} \sum_{i = 1}^N \log p(\vec{z}_i | \vec{\phi}) + \frac{1}{N} \log p(\vec{\phi}) + \text{const.}
\end{split}
\end{equation}
where `const.' denotes a $\vec{\phi}$-independent term that comes from the normalization of the posterior. In the large $N$ limit, the central limit theorem implies that $J'$ approaches its average value, which is
\begin{equation}
J'[\vec{\phi}] = \int_{\mathbb{R}^d} d\vec{z} \ q(\vec{z}) \  \log p(\vec{z} | \vec{\phi}) + \frac{1}{N} \log p(\vec{\phi})  \ ,
\end{equation}
where we have dropped the constant term since it does not affect our optimization. Our argument so far indicates that maximizing $J'$ with respect to $\vec{\phi}$ yields a Bayes-optimal estimate of it. The final remaining step, after which we will recover an objective equivalent to the one we study in the main text, is to simplify $J'$.

\paragraph{Taking the small-noise limit.} We can simplify $J'$ by using Laplace's method to approximate $p(\vec{z} | \vec{\phi})$ in the small noise ($\epsilon \to 0$) limit. The probability $p_{\text{noise}}(\vec{x} | \vec{z})$ is obviously maximized when $\vec{x} = \vec{\phi}(\vec{z})$; similarly, if $\vec{x} \in \vec{\phi}( \vec{R}^d)$, then $p(\vec{z} | \vec{x})$ is maximized at the unique $\vec{z}$ for which $\vec{\phi}(\vec{z}) = \vec{x}$ (this is unique since $\vec{\phi}$ is an injection).

Denote this unique point by $\vec{z}_*(\vec{x})$. At a small displacement $\Delta \vec{z}$ away from it,
\begin{equation}
\frac{1}{2} \Vert \vec{x} - \vec{\phi}(\vec{z}_*(\vec{x}) + \Delta \vec{z}) \Vert_2^2 \approx \frac{1}{2} (\Delta \vec{z})^T \vec{J}( \vec{z}_*(\vec{x}) )^T \vec{J}(\vec{z}_*(\vec{x})) (\Delta \vec{z}) \ ,
\end{equation}
i.e., the Hessian near $\vec{z}_*(\vec{x})$ is $\vec{J}( \vec{z}_*(\vec{x}) )^T \vec{J}(\vec{z}_*(\vec{x}))$. By Laplace's method, this implies
\begin{equation}
\int \frac{1}{(2 \pi \epsilon)^{D/2}} e^{- \frac{1}{2 \epsilon} \Vert \vec{x} - \vec{\phi}(\vec{z}') \Vert_2^2} \ q(\vec{z}') \ d\vec{z}' \xrightarrow{\epsilon \to 0} \frac{1}{(2 \pi \epsilon)^{(D-d)/2}}  \frac{1 }{\sqrt{ \det[ \vec{J}( \vec{z}_*(\vec{x}) )^T \vec{J}(\vec{z}_*(\vec{x}) ) ]  } } \ q(\vec{z}_*(\vec{x})) \ .
\end{equation}
Furthermore, applying Laplace's method again tells us that
\begin{equation}
\begin{split}
p(\vec{z} | \vec{\phi}) &\approx \frac{1}{(2 \pi \epsilon)^{d/2}} \int e^{- \frac{1}{2 \epsilon} \Vert \vec{x} - \vec{\phi}(\vec{z}) \Vert_2^2}  \ q(\vec{z}) \frac{\sqrt{ \det[ \vec{J}( \vec{z}_*(\vec{x}) )^T \vec{J}(\vec{z}_*(\vec{x}) ) ]  }}{q(\vec{z}_*(\vec{x}))} \ p_{\text{data}}(\vec{x})  \ d\vec{x} \\
&\approx   \sqrt{ \det[ \vec{J}( \vec{z} )^T \vec{J}(\vec{z} ) ]  } \ p_{\text{data}}(\vec{\phi}(\vec{z}))  
\end{split}
\end{equation}
in the $\epsilon \to 0$ limit. This gives us the result we want: maximizing the average log-posterior probability of $\vec{\phi}$ yields our objective, possibly plus an extra $\log p(\vec{\phi})$ term if the prior over $\vec{\phi}$ is non-uniform. More explicitly,
\begin{equation}
J'[\vec{\phi}] \xrightarrow{\epsilon \to 0} \int_{\mathbb{R}^d} d\vec{z} \ q(\vec{z}) \ \left\{ \ \frac{1}{2} \log \det[ \vec{J}( \vec{z} )^T \vec{J}(\vec{z} ) ]  + \log p_{\text{data}}(\vec{\phi}(\vec{z}))    \ \right\} + \frac{1}{N} \log p(\vec{\phi})  \ ,
\end{equation}
which only differs from Eq. \ref{eq:J_app} by a constant offset when $p(\vec{\phi})$ is uniform (or when we eliminate the prior term by taking $N \to \infty$).

\subsection{Additional comments about the objective}

\paragraph{The embedding map needs to be an injection.} The latent states $\vec{z}$ index locations on the $d$-dimensional manifold embedded in the ambient space, and hence function as coordinates for it. This intuition suggests that $\vec{\phi}$ ought to be an injection; any given point on the manifold should only correspond to a single value of $\vec{z}$. 

There is a more important technical reason for us to demand that $\vec{\phi}$ is an injection. If $\vec{\phi}$ is an injection, the pullback of $p_{\text{data}}$ through $\vec{\phi}$ is not generally normalized, i.e.,
\begin{equation}
\int_{\mathbb{R}^d} p_{\text{data}}(\vec{\phi}(\vec{z})) \sqrt{\vec{J}(\vec{z})^T \vec{J}(\vec{z})} \ d\vec{z} \leq 1 \ .
\end{equation}
But because it is less than or equal to one, we can still invoke Gibbs' inequality to argue
\begin{equation}
J[\vec{\phi}] = \int_{\mathbb{R}^d} q(\vec{z}) \log\left[ \frac{(\vec{\phi}^* p_{\text{data}})(\vec{z})}{q(\vec{z})} \right] \ d\vec{z} \leq 0 \ .
\end{equation}
(The argument is the same as the standard one used to prove that the KL divergence is nonnegative, so we omit it.) If $\vec{\phi}$ were \textit{not} an injection, $J$ does not necessarily have an upper bound, and hence our optimization problem is not well-defined. For example, suppose
\begin{equation}
\begin{split}
q(z) &:= 1 \ \text{ for } z \in [0, 1] \ \text{ and } \\
p_{\text{data}}(x, y) &:= \frac{1}{2 \pi} \delta( x^2 + y^2 - 1) \ \text{ for } (x, y) \in \mathbb{R}^2 ,
\end{split}
\end{equation}
i.e., $q$ is uniform on $[0, 1]$ and $p_{\text{data}}$ is uniform on the unit circle. Each embedding
\begin{equation}
\phi_1(z; k) := \cos( 2 \pi k z) \hspace{1in} \phi_2(z; k) := \sin(2 \pi k z)
\end{equation}
for $k \in \mathbb{N}$ yields the same $\log p_{\text{data}}(\vec{\phi}(z))$ term, but have different Jacobian terms. Note,
\begin{equation}
\frac{1}{2} \log \Vert \vec{\phi}'(z) \Vert_2^2 = \frac{1}{2} \log\left[ (2 \pi k)^2   \right] = \log(2 \pi k) \ ,
\end{equation}
which implies that we can make this term arbitrarily large by choosing $k$ to be larger and larger. This type of construction is only possible if we allow each $(x, y)$ on the unit circle to correspond to multiple values of $z$; such constructions are not possible if we insist that $\vec{\phi}$ is an injection, as Gibbs' inequality indicates. As an aside, we speculate that issues with enforcing the injectivity constraint are responsible for the `wiggles' depicted in \figureref{fig:first}b.

\paragraph{Boundary conditions.} We are looking for smooth injections $\vec{\phi}: \mathbb{R}^d \to \mathbb{R}^D$ that maximize Eq. \ref{eq:J_app} while also satisfying the finiteness conditions
\begin{equation} \label{eq:J_bc_app}
 \mathbb{E}_{\vec{z} \sim q}\left\{ \sqrt{\det(\vec{J}(\vec{z})^T \vec{J}(\vec{z}) )} \right\} < \infty  \hspace{0.2in} \text{ and } \hspace{0.2in}  \mathbb{E}_{\vec{z} \sim q}\left\{ \Vert \vec{\phi}(\vec{z}) \Vert^2_2 \right\} < \infty \ .
\end{equation}
Since a small region of $\mathbb{R}^d$ becomes a small region of $\mathbb{R}^D$ with surface area
\begin{equation}
dV = \sqrt{\det(\vec{J}(\vec{z})^T \vec{J}(\vec{z}) )} d\vec{z} 
\end{equation}
when projected through $\vec{\phi}$, the first condition is equivalent to asking that the average surface area (or the total $q$-weighted `mass' of the embedding) is finite. This condition prevents `needle'-like embeddings which project small regions of $\mathbb{R}^d$ to extremely large regions of $\mathbb{R}^D$. It also plays a crucial role in our argument that the PCA objective does not admit nonlinear solutions (see Appendix \ref{apd:pca_nonlinear}). The second condition prevents pathological embeddings in the tails of $q$ (i.e., large $\Vert \vec{z} \Vert_2$) by controlling their norm.

\clearpage

\section{Derivation of Euler-Lagrange equations}
\label{apd:EL}

In this appendix, we derive the Euler-Lagrange (EL) equations for our objective
\begin{equation} \label{eq:J_basic_app}
J[\vec{\phi}] = \int_{\mathbb{R}^d} d\vec{z} \ q(\vec{z}) \left[ \frac{1}{2} \log\det(\vec{J}(\vec{z})^T \vec{J}(\vec{z}) ) + \log p_{\text{data}}(\vec{\phi}(\vec{z}))  - \log q(\vec{z}) \right]  \ ,
\end{equation}
where $J_{ij} := \partial \phi_i/\partial z_j$ are the elements of the $D \times d$ Jacobian matrix $\vec{J}$. We will assume that $\vec{J}$ has full column rank, so that the determinant that appears is nonzero. The Lagrangian density corresponding to this objective is
\begin{equation}
\mathcal{L}(\vec{\phi}, \vec{J}, \vec{z}) =  q(\vec{z}) \left[ \frac{1}{2} \log\det(\vec{J}^T \vec{J}) + \log p_{\text{data}}(\vec{\phi}(\vec{z})) - \log q(\vec{z})  \right] \ .
\end{equation}

\subsection{Euler-Lagrange equations for arbitrary-dimensional embeddings}

The EL equations are
\begin{equation}
\sum_{j = 1}^d \partial_j \left( \frac{\partial \mathcal{L}}{\partial (\partial_j \phi_i)} \right) = \frac{\partial \mathcal{L}}{\partial \phi_i} \ ,
\end{equation}
so we only need to evaluate these derivatives explicitly and simplify the result. First, note
\begin{equation}
\frac{\partial \mathcal{L}}{\partial (\partial_j \phi_i)} = \frac{\partial \mathcal{L}}{\partial J_{ij}} = q(\vec{z}) \frac{\partial}{\partial J_{ij}} \left[  \frac{1}{2} \log\det(\vec{J}^T \vec{J}) \right] = q(\vec{z}) \left[  \vec{J} ( \vec{J}^T \vec{J} )^{-1}  \right]_{ij} \ ,
\end{equation}
where we have used a matrix cookbook result on the derivative of a log-determinant \citep{cookbook}. Note that $\vec{J}^+ :=  ( \vec{J}^T \vec{J} )^{-1} \vec{J}^T $ is the Moore-Penrose pseudoinverse of $\vec{J}$, i.e., the $d \times D$ matrix which satisfies $\vec{J}^+ \vec{J} = \vec{I}_d$. Using it allows us to write the above partial derivatives in vector form as
\begin{equation}
\frac{\partial \mathcal{L}}{\partial \vec{J}} = (\vec{J}^+)^T q(\vec{z}) \ .
\end{equation}
More straightforwardly, the partial derivative of $\mathcal{L}$ with respect to $\phi_i$ is the score of $p_{\text{data}}$:
\begin{equation}
\frac{\partial \mathcal{L}}{\partial \phi_i} = q(\vec{z}) \frac{\partial \log p_{\text{data}}(\vec{\phi}(\vec{z}))}{\partial \phi_i} \ .
\end{equation}
These results immediately imply that the EL equation for $\phi_i$ reads
\begin{equation}
\sum_j \partial_j \left( \frac{\partial \mathcal{L}}{\partial (\partial_j \phi_i)}  \right) = \frac{\partial \mathcal{L}}{\partial \phi_i} \ \implies \ \sum_j \partial_j \left[ J_{ji}^+ q(\vec{z}) \right] = q(\vec{z}) \frac{\partial \log p_{\text{data}}(\vec{\phi}(\vec{z}))}{\partial \phi_i} \ .
\end{equation}
Simplifying, we have
\begin{equation}
\sum_j \partial_j ( J_{ji}^+ ) + \sum_j J_{ji}^+ \frac{(\partial_j q)}{q} = \frac{\partial \log p_{\text{data}}(\vec{\phi}(\vec{z}))}{\partial \phi_i} \ ,
\end{equation}
or in vector form,
\begin{equation} \label{eq:EL_derived_app}
\left[ \vec{\partial} + \nabla \log q(\vec{z}) \right]^T \vec{J}^+ = \vec{s}_{\vec{\phi}}(\vec{z})^T \ ,
\end{equation}
where we have defined the score vector $s_i(\vec{z}) := \partial \log p_{\text{data}}(\vec{\phi}(\vec{z}))/\partial \phi_i$.

\subsection{Euler-Lagrange equations for one-dimensional embeddings}

The one-dimensional ($d = 1$) form of the objective reads
\begin{equation}
 J[\vec{\phi}] = \int_{-\infty}^{\infty} dz \ q(z) \left[ \frac{1}{2} \log \Vert \vec{\phi}'(z) \Vert_2^2 + \log p_{\text{data}}(\vec{\phi}(z)) - \log q(z)   \right] 
\end{equation}
and the corresponding Lagrangian is
\begin{equation}
L( \vec{\phi}, \vec{\phi}', z) := q(z) \left[ \frac{1}{2} \log \Vert \vec{\phi}'(z) \Vert_2^2 + \log p_{\text{data}}(\vec{\phi}(z)) - \log q(z)   \right] \ .
\end{equation}
In this setting, the EL equations have the somewhat simpler form
\begin{equation}
\frac{d}{dz} \left( \frac{\partial L}{\partial \phi'_i} \right) = \frac{\partial L}{\partial \phi_i} \hspace{0.2in} \text{, i.e.,} \hspace{0.2in} \frac{d}{dz} \left[  q(z) \frac{\phi_i'(z)}{\Vert \vec{\phi}'(z) \Vert_2^2} \right] = q(z) \frac{\partial}{\partial \phi_i} \log p_{\text{data}}(\vec{\phi}(z)) \ .
\end{equation}
Simplifying as in the previous subsection, we end up with
\begin{equation}
\frac{d}{dz} \left[ \frac{\phi_i'(z)}{\Vert \vec{\phi}'(z) \Vert_2^2} \right] + \frac{q'(z)}{q(z)} \frac{\phi_i'(z)}{\Vert \vec{\phi}'(z) \Vert_2^2} =  \frac{\partial}{\partial \phi_i} \log p_{\text{data}}(\vec{\phi}(z)) \ ,
\end{equation}
or if we prefer to be even more explicit,
\begin{equation}
\frac{\phi_i''(z)}{\Vert \vec{\phi}'(z) \Vert_2^2} - \frac{2 \phi_i'(z) \sum_k \phi_k'(z)}{\Vert \vec{\phi}'(z) \Vert_2^4} + \frac{q'(z)}{q(z)} \frac{\phi_i'(z)}{\Vert \vec{\phi}'(z) \Vert_2^2} =  \frac{\partial}{\partial \phi_i} \log p_{\text{data}}(\vec{\phi}(z)) \ .
\end{equation}

\clearpage

\section{Optimal 1D embeddings and diffusion models}
\label{apd:emb_1d}

In Sec. \ref{sec:1D}, we showed that optimal 1D embeddings $\vec{\phi}: \mathbb{R} \to \mathbb{R}^D$ satisfy a system 
\begin{equation}
\begin{split}
\phi_i'(z) = \frac{p_i(z)}{\Vert \vec{p}(z) \Vert_2^2} \hspace{0.5in} p_i'(z) = s_i(\vec{\phi}(z)) - \left[ \frac{d}{dz} \log q(z) \right] p_i(z) 
\end{split}
\end{equation}
of first-order ODEs, where the $p_i(z)$ are auxiliary momentum-like variables and $s_i(\vec{\phi}(z)) := \frac{\partial}{\partial \phi_i} \log p_{\text{data}}(\vec{\phi}(z))$ are the components of the \textit{score vector} associated with $p_{\text{data}}$ at the point $\vec{\phi}(z) \in \mathbb{R}^D$. In this appendix, we argue that optimal embeddings exactly follow score vectors in a certain limit.

Consider a prior $q(z) = \gamma e^{\gamma z}$ which is nonzero only for $z \in (-\infty, 0]$. For this $q$,
\begin{equation} \label{eq:ODE_first_specific}
\begin{split}
\phi_i'(z) = \frac{p_i(z)}{\Vert \vec{p}(z) \Vert_2^2} \hspace{0.5in} p_i'(z) = s_i(\vec{\phi}(z)) - \gamma p_i(z) \ .
\end{split}
\end{equation}
Note that $\gamma > 0$ controls the `time scale' on which $p_i(z)$ approaches its steady state value, at which $p_i(z) = s_i(\vec{\phi}(z))/\gamma$. More precisely, Eq. \ref{eq:ODE_first_specific} implies that $p_i(z)$ obeys the self-consistent equation
\begin{equation}
p_i(z) = \int_{-\infty}^z s_i(\vec{\phi}(z')) \ e^{- \gamma (z - z')} \ dz' \ .
\end{equation}
Since the exponential decay factor becomes a delta function in the large $\gamma$ limit, i.e.,
\begin{equation}
\gamma e^{- \gamma (z - z')} \xrightarrow{\gamma \to \infty} \delta(z - z') \ ,
\end{equation}
in this limit we have
\begin{equation}
p_i(z) \approx \frac{1}{\gamma} \int_{-\infty}^z s_i(\vec{\phi}(z')) \ \delta(z - z') \ dz' = \frac{1}{\gamma} s_i(\vec{\phi}(z))   \ ,
\end{equation}
which means that the momentum vectors match the score vectors. Eq. \ref{eq:ODE_first_specific} then implies
\begin{equation}
\phi_i'(z) \approx \gamma \frac{s_i(\vec{\phi}(z)) }{\Vert \vec{s}(z) \Vert_2^2} \ .
\end{equation}
As we noted in Sec. \ref{sec:1D}, this implies that trajectories follow score vectors, albeit with a speed that changes depending on their norm. If we perform a reparameterization of `time' with
\begin{equation}
\tau(z) := \int_{-\infty}^z \frac{1}{\Vert \vec{s}(z') \Vert_2^2} \ dz'    \hspace{0.2in} \tau'(z) = \frac{1}{\Vert \vec{s}(z) \Vert_2^2} \ , 
\end{equation}
then we have the equivalent dynamics
\begin{equation}
\phi_i'(\tau) = s_i(\vec{\phi}(\tau)) 
\end{equation}
which exactly follows the score field.

\clearpage

\section{The objective is reparameterization invariant}
\label{apd:basic_props}

Recall that our objective is
\begin{equation}
 J[\vec{\phi}] = \int_{\mathbb{R}^d} d\vec{z} \ q(\vec{z}) \left[ \frac{1}{2} \log\det[\vec{J}(\vec{z})^T \vec{J}(\vec{z})] + \log p_{\text{data}}(\vec{\phi}(\vec{z})) - \log q(\vec{z})   \right] \ .
\end{equation}
One of its key properties is that it is \textit{invariant to diffeomorphisms}; said differently, it is not sensitive to our choice of latent space coordinates. This property is desirable, since one's choice of coordinates is arbitrary, and different coordinate systems can reflect the same underlying geometry. In this appendix, we show that this property holds via a direct computation.

Let $\vec{g}: \mathbb{R}^d \to \mathbb{R}^d$ be a diffeomorphism, and consider a change of variables $\vec{z} = \vec{g}(\vec{z}')$ from $\vec{z}$ to $\vec{z}'$. Note that
\begin{equation}
\vec{\phi}(\vec{z}) = \vec{\phi}( \vec{g}(\vec{z}') ) \ ,
\end{equation}
so in the new coordinate system $\vec{\phi}$ becomes $\tilde{\vec{\phi}}(\vec{z}') := \vec{\phi}( \vec{g}(\vec{z}') )$. The Jacobian of $\tilde{\vec{\phi}}$ in this new coordinate system is
\begin{equation}
\tilde{J}_{ij} := \frac{\partial \tilde{\phi}_i}{\partial z_j'} = \sum_{k = 1}^d \frac{\partial \tilde{\phi}_i}{\partial z_k} \frac{\partial z_k}{\partial z_j'} = \sum_{k = 1}^d \frac{\partial \phi_i}{\partial z_k} \frac{\partial z_k}{\partial z_j'}  \ ,
\end{equation}
which means that $\tilde{\vec{J}} = \vec{J} \vec{J}_{\vec{g}}$, i.e., that the new Jacobian equals the old one times the Jacobian of $\vec{g}$. Finally, the measure $q(\vec{z}) d\vec{z}$ becomes
\begin{equation}
q( g(\vec{z}') ) | \det \vec{J}_{\vec{g}} | d\vec{z}' \ ,
\end{equation}
so in the new coordinate system we effectively go from $q(\vec{z})$ to $\tilde{q}(\vec{z}') := q( \vec{g}(\vec{z}') ) | \det \vec{J}_{\vec{g}} |$. Putting these results together, we find that $J[ \vec{\phi} ]$ equals
\begin{equation}
\begin{split}
J[\vec{\phi}] \ =& \ \int_{\mathbb{R}^d} d\vec{z}' \ q(\vec{g}(\vec{z}')) | \det \vec{J}_{\vec{g}} | \left[ \frac{1}{2} \log\det[\vec{J}(\vec{g}(\vec{z}'))^T \vec{J}(\vec{g}(\vec{z}'))] + \log p_{\text{data}}(\vec{\phi}(\vec{g}(\vec{z}'))) - \log q(\vec{g}(\vec{z}'))   \right] \\
=& \ \int_{\mathbb{R}^d} d\vec{z}' \ \tilde{q}(\vec{z}') \left[ \frac{1}{2} \log\det[\vec{J}(\vec{g}(\vec{z}'))^T \vec{J}(\vec{g}(\vec{z}'))] + \log p_{\text{data}}(\tilde{\vec{\phi}}(\vec{z}')) - \log q(\vec{g}(\vec{z}'))   \right] \\
=& \ \int_{\mathbb{R}^d} d\vec{z}' \ \tilde{q}(\vec{z}') \left[ \frac{1}{2} \log\det[\tilde{\vec{J}}(\vec{z}')^T \tilde{\vec{J}}(\vec{z}')] + \log p_{\text{data}}(\tilde{\vec{\phi}}(\vec{z}')) - \log \tilde{q}(\vec{z}')   \right] \ ,
\end{split}
\end{equation}
where two $\log \det | \vec{J}_{\vec{g}} |$ terms cancelled out in the last step. Since the final line is just the objective in the new coordinate system determined by the diffeomorphism $\vec{g}$, $J$ is indeed reparameterization invariant. 

Note that this property also follows straightforwardly from the fact that $J$ equals a (negative) KL divergence.

\clearpage

\section{Review and consequences of Noether's theorem}
\label{apd:noether}

Noether's theorem, which we first discuss in Sec. \ref{sec:useful_props} and then explore in more detail in Sec. \ref{sec:noether}, links continuous quasi-symmetries of the objective function to conservation laws \citep{Noether1918,Neuenschwander2017-wonderful,V2025noether}. In this appendix, we review its statement and provide a derivation of embedding energy conservation.

\subsection{Statement of Noether's theorem}

Let $\vec{\phi} := (\phi_1, ..., \phi_D)^T$ be a vector of $D$ fields defined on $\mathbb{R}^d$, i.e., each field is a smooth function from $\mathbb{R}^d$ to $\mathbb{R}$. Let
\begin{equation}
J[\vec{\phi}] := \int_{\mathbb{R}^d} d\vec{z} \ \mathcal{L}(\vec{\phi}, \vec{J}, \vec{z})
\end{equation}
be an objective function that depends on such a vector of fields through a Lagrangian density $\mathcal{L}(\vec{\phi}, \vec{J}, \vec{z})$, where the $D \times d$ Jacobian matrix $\vec{J}$ has components $J_{ij} = \partial \phi_i/\partial z_j$. 

We will call a field \textit{extremal} if it makes $J$ stationary, which happens if and only if it satisfies the EL equations
\begin{equation}
\sum_{j = 1}^d \partial_j \left( \frac{\partial \mathcal{L}}{\partial (\partial_j \phi_i)} \right) = \frac{\partial \mathcal{L}}{\partial \phi_i} \ .
\end{equation}
Next, define the \textit{stress-energy tensor}
\begin{equation}
T_{i j}(\vec{z}) := \sum_{k = 1}^D \frac{\partial \mathcal{L}}{\partial (\partial_i \phi_k)} (\partial_j \phi_k(\vec{z})) - \delta_{i j} \mathcal{L}(\vec{\phi}, \vec{J}, \vec{z}) \ .
\end{equation}
Following \cite{Neuenschwander2017-wonderful}, Noether's theorem is the following result.

\begin{theorem}[Noether's theorem]
Consider a transformation from coordinates $\vec{z}$ and fields $\vec{\phi}$ to
\begin{equation}
\begin{split}
\vec{z}' &= \vec{z} + \epsilon (\delta \vec{z})(\vec{z}, \vec{\phi}) \\
\vec{\phi}' &= \vec{\phi} + \epsilon (\delta \vec{\phi})(\vec{z}, \vec{\phi})
\end{split}
\end{equation}
where $\epsilon > 0$ is infinitesimally small. Note that the perturbations $\delta \vec{z}$ and $\delta \vec{\phi}$ are allowed to depend on $\vec{z}$ and $\vec{\phi}$. If this change of coordinates and fields changes the Lagrangian density by a total derivative, i.e.,
\begin{equation}
\mathcal{L}(\vec{\phi}', \vec{J}', \vec{z}') = \mathcal{L}(\vec{\phi}, \vec{J}, \vec{z}) + \epsilon \nabla \cdot \vec{K}(\vec{\phi}, \vec{J}, \vec{z}) + \mathcal{O}(\epsilon^2)
\end{equation}
for some function $\vec{K}$, then we call such a transformation a (continuous) quasi-symmetry of the objective, and conclude that the \textbf{Noether current}
\begin{equation}
j_a := \sum_{b = 1}^D \frac{\partial \mathcal{L}}{\partial (\partial_{a} \phi_b)} (\delta \phi_{b}) - \sum_{c = 1}^d T_{a c} (\delta z_c) - K_a
\end{equation}
is conserved (i.e., $\nabla \cdot \vec{j} = 0$) whenever $\vec{\phi}$ satisfies the EL equations.
\end{theorem}
For our \textit{particular} Lagrangian density
\begin{equation}
\mathcal{L}(\vec{\phi}, \vec{J}, \vec{z}) =  q(\vec{z}) \left[ \frac{1}{2} \log\det(\vec{J}^T \vec{J}) + \log p_{\text{data}}(\vec{\phi}(\vec{z})) - \log q(\vec{z})  \right] \ ,
\end{equation}
since (see Appendix \ref{apd:EL})
\begin{equation}
\frac{\partial \mathcal{L}}{\partial J_{ij}} = q(\vec{z}) J^+_{ji} \ ,
\end{equation}
we have
\begin{equation}
T_{i j}(\vec{z}) := \sum_k q(\vec{z}) J^+_{i k} J_{k j} - \delta_{i j} \mathcal{L}(\vec{\phi}, \vec{J}, \vec{z})  = \delta_{i j} \left[ q(\vec{z}) -  \mathcal{L}(\vec{\phi}, \vec{J}, \vec{z}) \right]
\end{equation}
and
\begin{equation}
\begin{split}
j_a &= \sum_{b = 1}^D  q(\vec{z}) J^+_{a b} (\delta \phi_{b}) - \sum_{c = 1}^d \delta_{a c} (q - \mathcal{L}) (\delta z_b) - K_a \\
&= \sum_{b = 1}^D  q(\vec{z}) J^+_{a b} (\delta \phi_{b}) - (q(\vec{z}) - \mathcal{L}) (\delta z_a) - K_a \ .
\end{split}
\end{equation}

\subsection{Direct derivation of embedding energy conservation}
\label{sec:embed_direct}

We can derive embedding energy conservation through a direct computation, without going through Noether's theorem. We want to show that the quantity
\begin{equation}
\boxed{ \ \mathcal{E} := -\frac{1}{2} \log\det(\vec{J}^T \vec{J}) - \log p_{\text{data}}(\vec{\phi}(\vec{z})) + \log q(\vec{z}) \ } \ ,
\end{equation}
which we call the \textbf{embedding energy}, is constant (i.e., does not depend on $\vec{z}$) for all solutions of the EL equations. To see this, multiply Eq. \ref{eq:EL_derived_app} on the right by $\vec{J}$:
\begin{equation} 
\left[ \vec{\partial} + \nabla \log q(\vec{z}) \right]^T \vec{J}^+ \vec{J} = \vec{s}_{\vec{\phi}}(\vec{z})^T \vec{J} \ .
\end{equation}
Since $\vec{J}^+ \vec{J} = \vec{I}$, one term is easy to simplify:
\begin{equation}
\left[ \nabla \log q(\vec{z}) \right]^T \vec{J}^+ \vec{J} = \left[ \nabla \log q(\vec{z}) \right]^T \ .
\end{equation}
The derivative term is somewhat more delicate. In terms of components, we have
\begin{equation}
\sum_{j, i} \partial_j ( J^+_{j i} ) J_{i k} = \sum_{j, i} \partial_j ( J^+_{j i}  J_{i k} ) - J^+_{ji} \partial_j ( J_{i k} ) = - \sum_{j, i} J^+_{ji} \partial_j ( J_{i k} ) \ .
\end{equation}
But note that, since $(\vec{J}^+)^T$ is the derivative of the kinetic term with respect to $\vec{J}$, we have
\begin{equation*}
\partial_k \left[  \frac{1}{2} \log\det(\vec{J}^T \vec{J}) \right] = \sum_{i, j} \frac{\partial}{\partial J_{ij}} \left[  \frac{1}{2} \log\det(\vec{J}^T \vec{J}) \right] \partial_k ( J_{ij} ) = \sum_{i, j} J^+_{j i} \partial_k ( J_{ij} ) = \sum_{i, j} J^+_{j i} \partial_j ( J_{ik} ) \ ,
\end{equation*}
where in the last step we used the fact that $\partial_k ( J_{ij} ) = \partial^2_{k j} \phi_i = \partial^2_{j k} \phi_i = \partial_j ( J_{i k})$. Hence, we conclude that
\begin{equation} 
 \nabla \left[  \frac{1}{2} \log\det(\vec{J}^T \vec{J})   \right] + \nabla \log q(\vec{z}) = \vec{s}_{\vec{\phi}}(\vec{z})^T \vec{J}  = \nabla \log p_{\text{data}}(\vec{\phi}(\vec{z})) \ ,
\end{equation}
or equivalently that
\begin{equation}
\nabla \left\{  \frac{1}{2} \log\det(\vec{J}^T \vec{J})  +   \log p_{\text{data}}(\vec{\phi}(\vec{z})) - \log q(\vec{z})   \right\} = 0 \ .
\end{equation}
This implies that the bracketed quantity is constant, i.e., that $\mathcal{E}$ is conserved for solutions of the EL equations.

As an aside, note that
\begin{equation}
0 \geq J[\vec{\phi}] = - \int_{\mathbb{R}^d} d\vec{z} \ q(\vec{z}) \ \mathcal{E} = - \mathcal{E} \ .
\end{equation}
This implies that we ought to minimize $\mathcal{E}$, and that if $J$ is perfectly maximized (such that the corresponding KL divergence is minimized), $\mathcal{E} = 0$.

\clearpage

\section{There are no nonlinear solutions to the PCA objective}
\label{apd:pca_nonlinear}

In the main text (Sec. \ref{sec:main_pca}), we argue that linear maps satisfy the EL equations associated with the PCA objective, and find that the linear maps that maximize $J$ pick out the top eigenvectors of the data's covariance matrix. Here, we argue that it is impossible for these EL equations to have nonlinear solutions, and hence for $J$ to have any nonlinear global maximizers. This point is important, since otherwise we have not established that linear maps are the global maximizers of the PCA objective.

For technical reasons,\footnote{For arbitrary $\sigma_0$, we must consider Hermite polynomials with scaled arguments, which adds a small amount of notational clutter. Aside from this issue, the argument would be exactly the same.} we specialize to the prior with $\sigma_0 = 1$ below, but this assumption can easily be relaxed.

\subsection{The one-dimensional case}

The objective is
\begin{equation} 
J[\vec{\phi}] := \int_{-\infty}^{\infty} dz \ \frac{e^{- \frac{1}{2} z^2}}{\sqrt{2 \pi }} \ \left\{  \frac{1}{2} \log \Vert \vec{\phi}'(z) \Vert_2^2 - \frac{1}{2} \vec{\phi}(z)^T \vec{\Sigma}^{-1} \vec{\phi}(z)    \right\} + \text{const.}
\end{equation}
where we are optimizing over smooth injections $\vec{\phi}: \mathbb{R} \to \mathbb{R}^D$. Recall from Sec. \ref{sec:obj} that we also insist on two finiteness conditions, which here read
\begin{equation}
\mathbb{E}_{z \sim q}\left\{  \Vert \vec{\phi}'(z) \Vert_2   \right\} < \infty \hspace{0.2in} \text{ and } \hspace{0.2in} \mathbb{E}_{z \sim q}\left\{  \Vert \vec{\phi} \Vert_2^2 \right\} < \infty \ .
\end{equation}
Note that the second condition implies that
\begin{equation}
 \mathbb{E}_{z \sim q}\left\{   \phi_i(z)^2 \right\} < \infty 
\end{equation}
for each $i = 1, ..., D$. This, in turn, implies that each $\phi_i(z)$ can be written in terms of a (probabilist's) Hermite polynomial expansion
\begin{equation}
\phi_i(z) = \sum_{n = 0}^{\infty} c_{n i} H_n( z ) = c_{0 i} + c_{1 i} z + c_{2 i} (z^2 - 1) + \cdots \ .
\end{equation}
By the conservation of embedding energy (see Sec. \ref{sec:noether}), we have
\begin{equation}
-\frac{1}{2} \Vert \vec{\phi}'(z) \Vert_2^2 + \frac{1}{2} \vec{\phi}(z)^T \vec{\Sigma}^{-1} \vec{\phi}(z)  - \frac{z^2}{2} = \text{const.} 
\end{equation}
which implies that
\begin{equation}
\begin{split}
\Vert \vec{\phi}'(z) \Vert_2 &= C \exp\left\{ \frac{1}{2} \vec{\phi}(z)^T \vec{\Sigma}^{-1} \vec{\phi}(z)  - \frac{z^2}{2}  \right\} \\
&= C \exp\left\{ \frac{(\vec{c}_0^T  \vec{\Sigma}^{-1} \vec{c}_0) }{2}  + (\vec{c}_0^T  \vec{\Sigma}^{-1} \vec{c}_1) z + \frac{(\vec{c}_1^T  \vec{\Sigma}^{-1} \vec{c}_1) }{2} z^2 + \mathcal{O}( z^3)   - \frac{z^2}{2}  \right\} 
\end{split}
\end{equation}
for some constant $C > 0$, where $\vec{c}_n$ denotes the vector whose $i$-th component is $c_{n i}$. Unless the higher-order coefficients are zero, this norm will grow at least as quickly as $\exp( z^n)$ for some $n > 2$, and hence the first finiteness condition will be violated.

\subsection{The general case}

In the general case,
\begin{equation}
J[\vec{\phi}] := \int_{\mathbb{R}^d} d\vec{z} \ \frac{e^{- \frac{1}{2 } \Vert \vec{z} \Vert_2^2}}{(2 \pi )^{d/2}} \ \left\{ \frac{1}{2} \log\det[ \vec{J}(\vec{z})^T \vec{J}(\vec{z})] - \frac{1}{2} \vec{\phi}(\vec{z})^T \vec{\Sigma}^{-1} \vec{\phi}(\vec{z})  \right\} + \text{const.}
\end{equation}
where we are optimizing over smooth injections $\vec{\phi}: \mathbb{R}^d \to \mathbb{R}^D$. The general finiteness conditions are
\begin{equation} 
 \mathbb{E}_{\vec{z} \sim q}\left\{ \sqrt{\det(\vec{J}(\vec{z})^T \vec{J}(\vec{z}) )} \right\} < \infty  \hspace{0.2in} \text{ and } \hspace{0.2in}  \mathbb{E}_{\vec{z} \sim q}\left\{ \Vert \vec{\phi}(\vec{z}) \Vert_2^2 \right\} < \infty  \ .
\end{equation}
As above, the second finiteness condition implies that each $\phi_i(z)$ can be written in terms of a Hermite polynomial expansion. By the conservation of embedding energy,
\begin{equation}
- \log \sqrt{\det(\vec{J}(\vec{z})^T \vec{J}(\vec{z}))} + \frac{1}{2} \vec{\phi}(\vec{z})^T \vec{\Sigma}^{-1} \vec{\phi}(\vec{z})  - \frac{\Vert \vec{z} \Vert_2^2}{2} = \text{const.} 
\end{equation}
which implies
\begin{equation}
\begin{split}
\sqrt{\det(\vec{J}(\vec{z})^T \vec{J}(\vec{z}))} = C \exp\left\{ \frac{1}{2} \vec{\phi}(\vec{z})^T \vec{\Sigma}^{-1} \vec{\phi}(\vec{z})  - \frac{\Vert \vec{z} \Vert_2^2}{2}  \right\}
\end{split}
\end{equation}
for some constant $C > 0$. By the same argument as before, the argument of this exponential will grow at least as quickly as a polynomial in $\vec{z}$, and hence not satisfy the first finiteness condition, unless the coefficients of the higher-order Hermite expansion terms are zero. Since only the zeroth and first order terms of the expansion are permitted, we conclude that the global maximizers of $J$ must be linear.

\end{document}